\documentclass[10pt,journal,compsoc]{IEEEtran}

\ifCLASSOPTIONcompsoc
  \usepackage[nocompress]{cite}
\else
  \usepackage{cite}
\fi

\ifCLASSINFOpdf
\else
\fi

\usepackage{dblfloatfix}
\usepackage{booktabs}
\usepackage{graphicx}

\usepackage{color}

\usepackage{xcolor}

\usepackage{makecell}

\usepackage[hyphens]{url}
\usepackage[hidelinks, pagebackref=true,breaklinks=true,colorlinks,bookmarks=false]{hyperref}
\hypersetup{breaklinks=true}
\begin{document}
%
\title{VIPriors 4: Visual Inductive Priors for\\Data-Efficient Deep Learning Challenges}
%
%
%
%

\author{Robert-Jan~Bruintjes,
        Attila~Lengyel,
        Osman~Semih~Kayhan,
        Nergis~Tomen,
        Hadi~Jamali-Rad,
        and~Jan~van~Gemert
\IEEEcompsocitemizethanks{
\IEEEcompsocthanksitem R. Bruintjes, A. Lengyel, N. Tomen, H. Jamali-Rad and J. van Gemert are with Delft University of Technology.\protect\\
E-mail: r.bruintjes@tudelft.nl
\IEEEcompsocthanksitem O. S. Kayhan is with Bosch Security Systems B.V.
\IEEEcompsocthanksitem H. Jamali-Rad is with Shell.
}
}

%
%

\IEEEtitleabstractindextext{%
\begin{abstract}

The fourth edition of the "VIPriors: Visual Inductive Priors for Data-Efficient Deep Learning" workshop features two data-impaired challenges. These challenges address the problem of training deep learning models for computer vision tasks with limited data. Participants are limited to training models from scratch using a low number of training samples and are not allowed to use any form of transfer learning. We aim to stimulate the development of novel approaches that incorporate inductive biases to improve the data efficiency of deep learning models. Significant advancements are made compared to the provided baselines, where winning solutions surpass the baselines by a considerable margin in both tasks. As in previous editions, these achievements are primarily attributed to heavy use of data augmentation policies and large model ensembles, though novel prior-based methods seem to contribute more to successful solutions compared to last year. This report highlights the key aspects of the challenges and their outcomes.

\end{abstract}

}

\maketitle

\IEEEdisplaynontitleabstractindextext

%
\IEEEpeerreviewmaketitle


%
%
%
%

\IEEEraisesectionheading{\section{Introduction}\label{sec:introduction}}


\IEEEPARstart{D}{eep} learning is increasingly powered by large training datasets. However, collecting such datasets is often costly.
Progress in recent years has shown the successful application of large quantities of data to train comprehensive foundation models for vision and language~\cite{Bommasani2021OnTO}, and to combine multiple modalities for weak supervision~\cite{radford21learning}.

However, training with massive datasets still requires a significant amount of energy, contributing to carbon emissions. Furthermore, access to datasets and compute at such scale is limited to a few powerful deep learning behemoths. Additionally, such amounts of data may not be available for some domains. The Visual Inductive Priors for Data-Efficient Deep Learning workshop (VIPriors) therefore encourages research in learning from small datasets, by way of combining the learning power of deep learning with hard-won prior knowledge from specific domains. 

The Visual Inductive Priors for Data-Efficient Deep Learning workshop has now been organized for the fourth year in a row \cite{bruintjes2021vipriors,lengyel2022vipriors,bruintjes2023vipriors}, with the latest edition taking place at ICCV 2023 in Paris, France.
The workshop features a paper track as well as challenges, where participants train computer vision models on small subsets of (publicly available) datasets, challenging them to find competitive solutions without the large quantities of data that power state-of-the-art deep computer vision models.

In this report, we present the outcomes of the fourth edition of the VIPriors challenges. 
This edition features an object detection challenge as well as an instance segmentation challenge. We discuss top-ranking solutions of both challenges, as well as the submission that receives the jury prize for introducing innovative methods.
We observe that, even though some novel prior-based methods seem to contribute to success in several solutions, heavy use of model ensembling and data augmentation dominates successful solutions, like in previous editions.
\section{Challenges}
\label{sec:challenges}
The workshop hosts two computer vision challenges in which the number of training samples are reduced to a small fraction of the full set:


 \textbf{Object detection}: The DelftBikes~\cite{kayhan2021hallucination} dataset is used for the object detection challenge. Each image contains 22 different bike parts that are annotated with bounding box, class and object state labels.

\textbf{Instance segmentation}: The main objective of this challenge is to segment basketball players and the ball on images recorded of a basketball court. The dataset is provided by SynergySports\footnote{\url{https://synergysports.com}} and contains a train, validation and test set of basketball games recorded at different courts with instance labels.


We provide a toolkit\footnote{\url{https://github.com/VIPriors/vipriors-challenges-toolkit}} with guidelines, baseline models and datasets for each challenge.
The competitions are hosted on Codalab~\cite{codalab_competitions_JMLR}. Each participating team submits their predictions computed over a test set of samples for which labels are withheld from competitors.

The challenges include certain rules to follow:
\begin{itemize}
    \item Models shall be trained from scratch with only the given dataset.
    \item The usage of other data rather than the provided training data, such as pretraining the models on other data and transfer learning methods, are prohibited. It is however allowed to train with synthetic data generated from the training data.
    \item The participating teams are required to write a technical report about their methodology and experiments. We cite from these reports.
\end{itemize}
\subsection{Object Detection}
\begin{figure}[t]
	\centering
	\includegraphics[width=\linewidth]{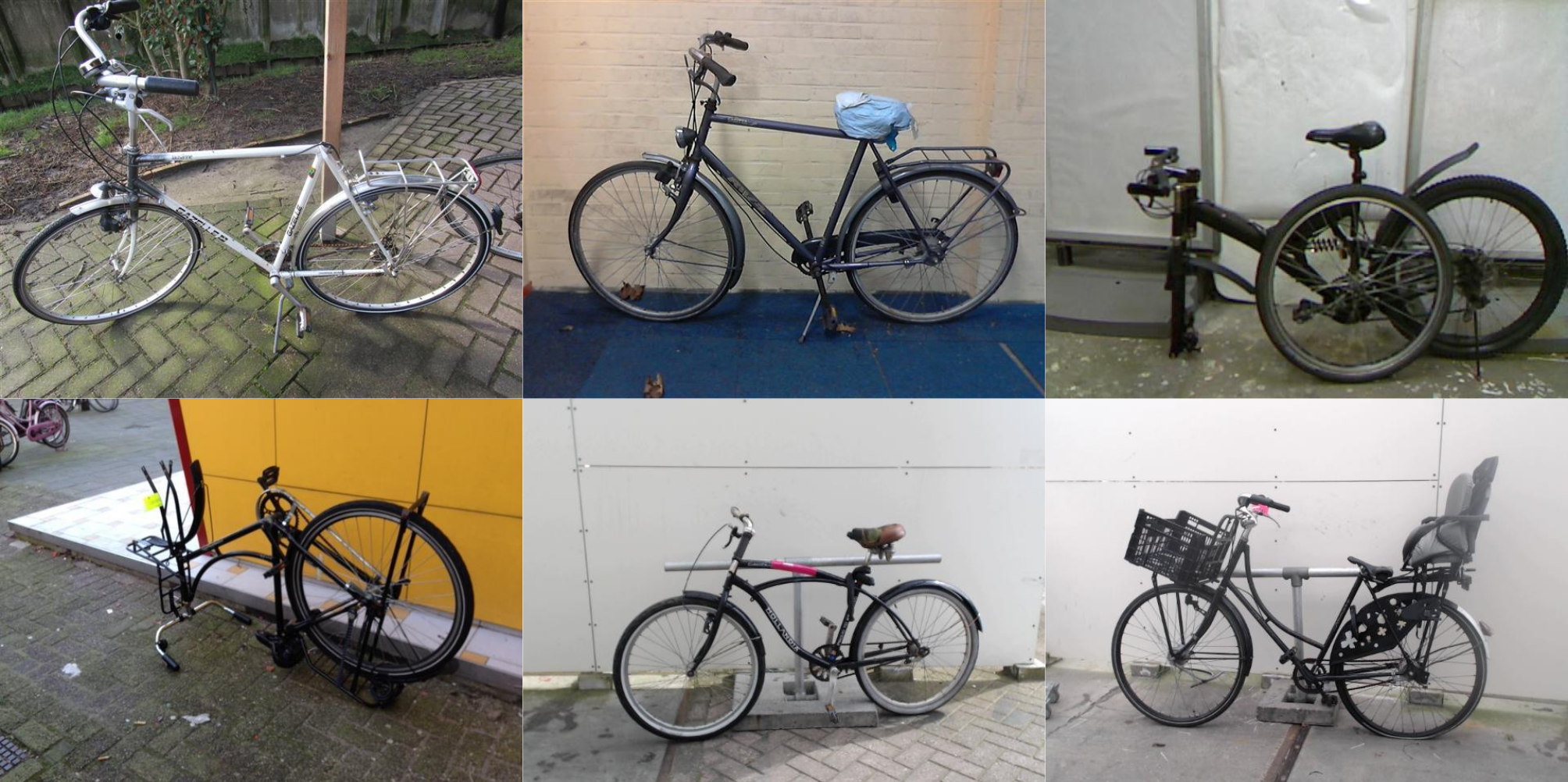}	
	\caption{Some images from the DelftBikes dataset. Each image has a single bike with 22 labeled parts. }
	\label{fig:bike_images}
\end{figure}

The object detection challenge uses the DelftBikes~\cite{kayhan2021hallucination} dataset. The dataset includes 8,000 bike images for training and 2,000 images for testing (Fig.~\ref{fig:bike_images}). Images contain 22 bike parts, labeled by class, bounding box and part state labels such as intact, missing, broken or occluded.
The dataset contains varying object sizes, and contextual and location biases that can cause false positive detections~\cite{kayhan2021hallucination, kayhan2022evaluating}. Note that some of the object boxes are noisy which introduces more challenges to detect object parts.

As a baseline detector we use the same model as used in the 2021 and 2022 challenges \cite{lengyel2022vipriors,bruintjes2023vipriors}, which is a Faster RCNN model with a Resnet-50 FPN~\cite{ren2016faster} backbone trained from scratch for 16 epochs. This baseline network is trained with the original image size without any data augmentation. It attains 25.8\% AP score on the test set.
Note that the evaluation is done on available parts which are intact, damaged and occluded parts.


\begin{table*}[t]
\centering
\caption{Final rankings of the Object Detection challenge.}
\renewcommand{\arraystretch}{2.0}
\begin{tabular}{@{}llc@{}}
\toprule
Ranking & Teams & AP @ 0.5:0.95 \\ \midrule
1 & \makecell[l]{\textbf{Jiawei Zhao, Xuede Li, Xingyue Chen, Junfeng Luo.} \\ \textbf{\textit{Vision Intelligence Department (VID), Meituan.}}} & \textbf{34.5} \\
2 \& J & \makecell[l]{Xiaoqiang Lu, Jiaxuan Zhao, Yuting Yang, Zhongjian Huang, Xu Liu, Fang Liu, Licheng Jiao. \\ \textit{School of Artificial Intelligence, Xidian University.}} & 33.3 \\
3 & \makecell[l]{Zhang Jing, Qinliang Wang, Shizhan Zhao. \\ \textit{School of Artificial Intelligence, Xidian University.}}  & 30.6 \\
4 & \makecell[l]{Zheng Wang, Dong Xie, Hanzhi Wang, Jiang Tian. \\ \textit{AI Lab, Lenovo Research} \\ \textit{State Key Laboratory of Virtual Reality Technology and Systems, Beihang University} \\ \textit{Department of Computer Science, Yale University}}  & 30.4 \\
5 & \makecell[l]{Xinyu Sun, Xiaoyu Hao. \\ \textit{Xidian University}}  & 29.4 \\
6 & \makecell[l]{\textit{Team fha.ddd}} & 26.6 \\
 
\bottomrule
\end{tabular}
\label{tab:detection2}
\end{table*}

The detection challenge has six participating teams. The team from Vision Intelligence Department of Meituan obtains first place with a 34.5\% AP score. Two teams from Xidian University follow them by 33.3\% AP and 30.6\% AP respectively.
The team from Xidian University lead by Xiaoqiang Lu wins the jury prize.
The final rankings are shown in Table \ref{tab:detection2}.


\subsubsection{First place}
\begin{figure}[t]
	\centering
	\includegraphics[width=\linewidth]{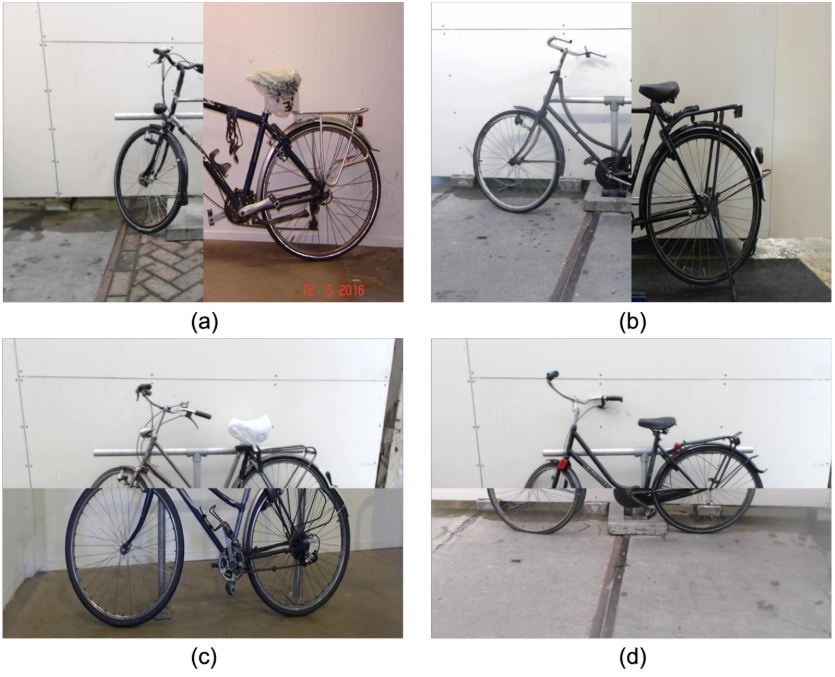}	
	\caption{The first place solution of Zhao et al. recombines binary pairs of images horizontally and vertically to create a synthetic pre-training dataset. Figure adapted from technical report by Zhao et al. provided to competition organizers.}
	\label{fig:arch11}
\end{figure}

Zhao et al. use a Cascade RNN \cite{cai2018cascade} with a ConvNeXtV2 backbone \cite{woo2023convnext}. For data augmentation they use the Albumentations library \cite{info11020125} to apply Albu, PhotoMetricDistortion, MixUp \cite{zhang2018mixup}, and Auto Augment V2 \cite{cubuk2018autoaugment}. They pretrain their model on a synthetic dataset created from horizontal and vertical recombinations of binary pairs of samples (see Fig.~\ref{fig:arch11}). They finally apply SWA \cite{Izmailov2019averaging} and retrain the model on manually identified hard classes.

\subsubsection{Second place and jury prize}

\begin{figure}[t]
	\centering
	\includegraphics[width=\linewidth]{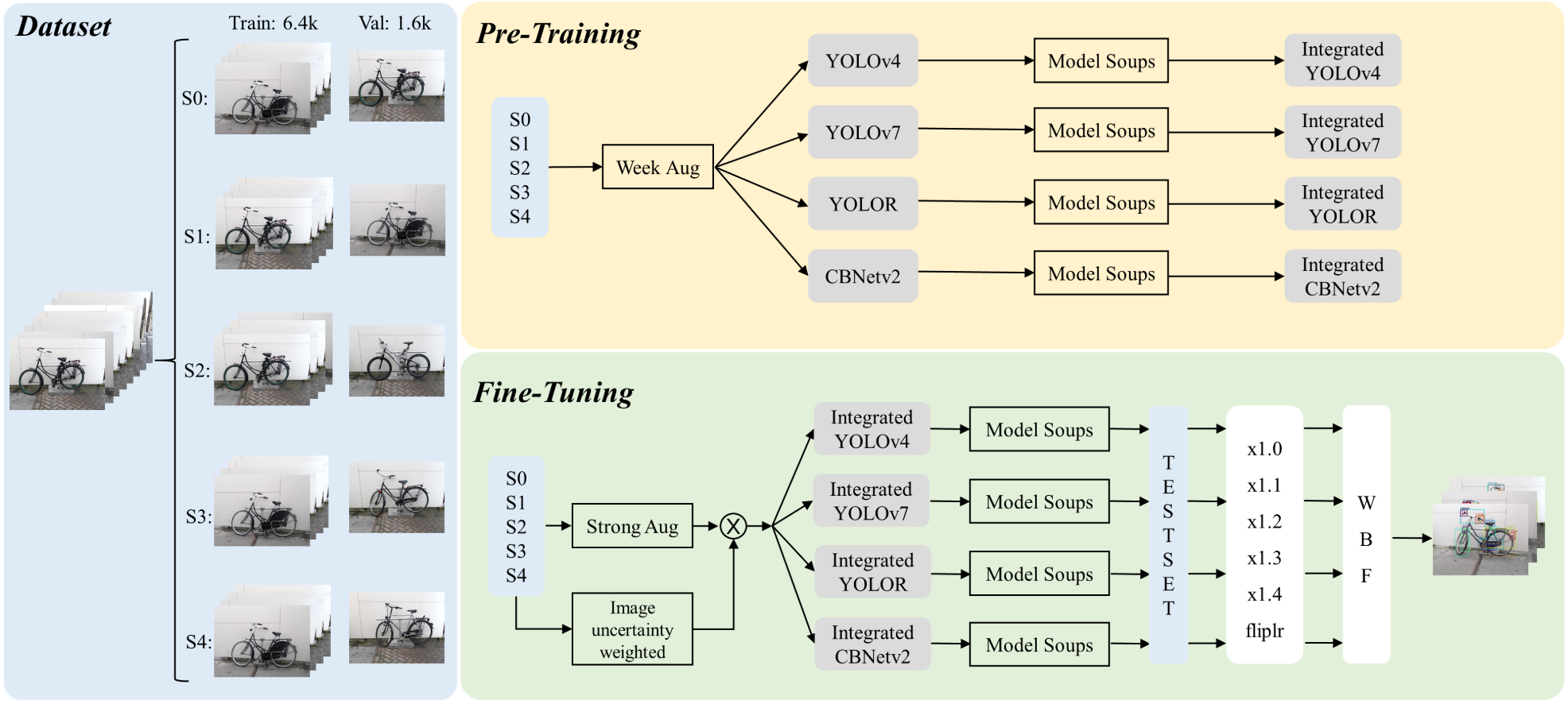}	
	\caption{Overview of the training pipeline of the second place solution and jury prize winner of Lu et al.}
	\label{fig:lu-fig1}
\end{figure}

Lu et al. use Scaled-YOLOv4~\cite{wang2021scaled}, YOLOv7~\cite{wang2023yolov7}, YOLOR~\cite{wang2021you} and CBNetv2~\cite{cbnetv2} backbones. These backbones are trained using cross-validation and light data augmentation (random scaling, random flipping, color jitter) where all model instances of the same backbone are then combined using Model Soups~\cite{wortsman2022model}. The resulting models are used as pre-training weights in the final training scheme, where once again models are trained with cross-validation, as well as stronger data augmentation (Mosaic Augmentation~\cite{bochkovskiy2020yolov4}, ~\cite{kisantal2019augmentation}, Mix-Up~\cite{zhang2018mixup}, Cutout~\cite{devries2017improved}) and a custom method called \textit{Image Uncertainty Weighted} where images with less certain box scores are weighted heavier in the loss function. After applying Model Soups to all model instances, Test-Time Augmentation~\cite{moshkov2020test} and Weighted Boxes Fusion~\cite{solovyev2021weighted} are applied to achieve the final model. The full pipeline is shown in Fig.~\ref{fig:lu-fig1}.

The jury prize is awarded to Lu et al. for their proposed novel method \textit{Image Uncertainty Weighted} which addresses the specific dataset prior knowledge of a some classes being underrepresented in the data.

\subsubsection{Third place}

Jing et al. initially train five backbones on training size 1280px with a confidence threshold of 0.001: YOLOv7~\cite{yolov7}, YOLOv8x-p2\cite{githubGitHubUltralyticsultralytics}, YOLOv8x, YOLOv8x-p6 and Cascade RCNN~\cite{cai2018cascade}, while employing Mosaic Augmentation~\cite{bochkovskiy2020yolov4}, MixUp~\cite{zhang2018mixup}, Test-Time Augmentation~\cite{moshkov2020test} and horizontal flip testing. Finally, the models are combined using Weighted Boxes Fusion~\cite{solovyev2021weighted}.
\subsection{Instance Segmentation}
In the task of instance segmentation, one detects and segments instances of objects in an image. Instance segmentation is a popular and widely applicable computer vision problem, with applications ranging from autonomous driving, surveillance, remote sensing to sport analysis. Similarly to the 2021 and 2022 editions of this challenge \cite{lengyel2022vipriors,bruintjes2023vipriors}, our challenge is based on the basketball dataset provided by SynergySports \cite{sportradar_dataset}, consisting of images recorded during various basketball games played on different courts. The goal is to detect and predict segmentation masks of all players and ball objects in the images. With a mere 184, 62, and 64 samples for the train, validation and test splits, respectively, the dataset is considered very small. The test labels are withheld from the challenge participants and final performance on the test set is evaluated on an online server. The main metric used is the Average Precision (AP) @ 0.50:0.95. As in last year's challenge \cite{bruintjes2023vipriors}, our baseline method is based on the Detectron2 \cite{wu2019detectron2} implementation of Mask-RCNN \cite{he2017mask}.

Forty-six teams submitted solutions to the evaluation server, of which four teams submitted a report to qualify their submission to the challenge. 
Two teams from Xidian University obtains first and second place with 59\% AP and 58.2\% AP respectively. The team from Sichuan University follows them with 55.2\% AP.
The team from National Cheng Kung University wins the jury prize.
The final rankings are shown in Table \ref{tab:segmentation}.

\begin{table*}[t]
\centering
\caption{Final rankings of the Instance Segmentation challenge. J indicates jury prize.}
\renewcommand{\arraystretch}{2.0}
\begin{tabular}[t]{@{}llc@{}}
\toprule
Ranking & Teams & \% AP @ 0.50:0.95 \\ \midrule

1 & \makecell[l]{
    \textbf{Junpei Zhang, Kexin Zhang, Rui Peng, Licheng Jiao, Fang Liu, Lingling Li, Yuting Yang.} \\
    \textbf{\textit{Xidian University, Xi’an, Shaanxi.}}
    } & \textbf{59.0} \\
2 & \makecell[l]{
    Xiaoqiang Lu, Yuting Yang, Zhongjian Huang, Jiaxuan Zhao, Xu Liu, Fang Liu, Licheng Jiao. \\
    \textit{School of Artificial Intelligence, Xidian University.}} & 58.2 \\
    
3 & \makecell[l]{
    Huijia Liang, Jin Yang. \\
    \textit{Sichuan University.}} & 55.2 \\ 

4 \& J & \makecell[l]{
    Chih-Chung Hsu, Chia-Ming Lee, Ming-Shyen Wu. \\
    \textit{Institute of Data Science, National Cheng Kung University.}} & 50.9 \\ 
    
    \bottomrule
\end{tabular}
\label{tab:segmentation}
\end{table*}


\subsubsection{First place}
Zhang et al. achieve first place using a novel method called \textit{Orthogonal Uncertainty Representation} (OUR), which ensures that the observed geometric manifold~\cite{ma2023curvature} of underrepresented classes is broadened during training. The model trained is a Mask RCNN~\cite{he2017mask} model with Swin~\cite{liu2021Swin}, ResNet~\cite{he2016deep} and CBNet~\cite{liu2020cbnet} backbones, which is trained with geometric~\cite{paschali2019manifold}, color space, sharpness, noise injection and Copy-Paste~\cite{kisantal2019augmentation} data augmentations. The segmentation head is a Hybrid Task Cascade (HTC)~\cite{Chen_2019_CVPR} head. The loss function is a Seesaw loss~\cite{Wang_2021_CVPR}. Predicted boxes are fused using Weighted Boxes Fusion~\cite{solovyev2021weighted}. Models are ensembled using SWALP~\cite{yang2019swalp}.

Incorporating the novel OUR method increased the performance of the trained model by 1.1\% AP. Referencing this performance increase against the challenge standings it is apparent that OUR contributes significantly to Zhang et al. achieving first place in the challenge.

\subsubsection{Second place}
\begin{figure}[t]
    \centering
    \includegraphics[width=\linewidth]{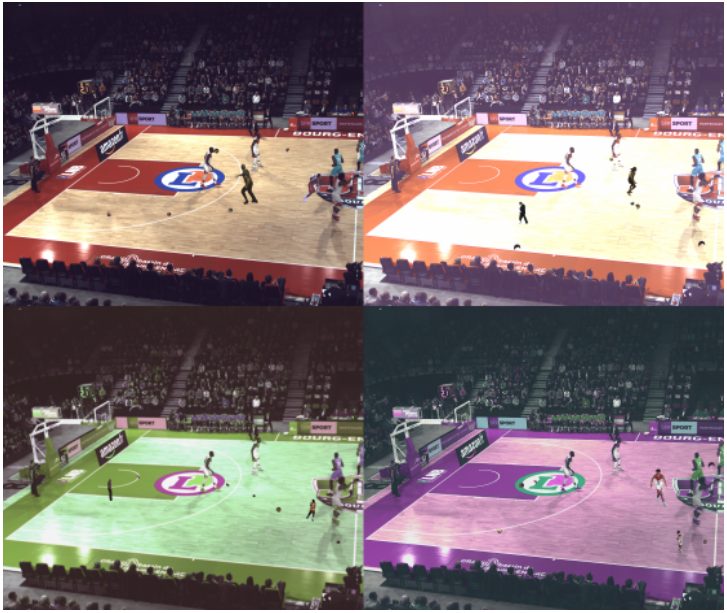}
    \caption{Example data augmentation output of Lu et al. Figure adapted from technical report by Lu et al. provided to competition organizers.}
    \label{fig:Lu2-fig1}
\end{figure}

Lu et al. achieve second place by building off last year's entry by Yusunov et al.~\cite{yunusov2021instance}. This model is an HTC~\cite{chen2019hybrid} model with BEiTv2-L~\cite{peng2022unified} with ViT-Adapter~\cite{chen2022vision} and Internimage-XL~\cite{wang2023internimage} backbones. The loss function is a GIoU loss~\cite{rezatofighi2019generalized} and they use Soft NMS~\cite{bodla2017soft}. Lu et al. use extensive data augmentation: Mosaic Augmentation~\cite{bochkovskiy2020yolov4}, Copy-Paste~\cite{kisantal2019augmentation}, Mix-Up~\cite{zhang2018mixup}, random brightness, random contrast, random saturation, random scale, random flip, sharpen and overlay, blur, Gaussian noise and grid-mask. An example data augmentation output is shown in Fig.~\ref{fig:Lu2-fig1}. The model is trained with AdamW~\cite{loshchilov2017decoupled}. Two expert networks are used to refine mask output: SegFormer~\cite{xie2021segformer} and SeMask~\cite{jain2023semask}. Models are ensembled using Model Soups~\cite{wortsman2022model}. The final model is trained on a combination of the provided training and validation sets.

Interestingly, Lu et al. note that "integrating other detectors’ prediction under different backbones" was necessary to achieve the final test set score of 58.2\% AP. Other significant contributors to the final AP score are the data augmentation methods Copy-Paste and Mosaic, which together contribute 1.4\% AP improvement to the final model.

\subsubsection{Third place}
Liang et al. achieve third place using a ensemble of three models: Mask2Former~\cite{cheng2022masked}, YOLOX~\cite{ge2021yolox}, and DETR~\cite{carion2020end}, fused using Weighted Boxes Fusion~\cite{solovyev2021weighted}. Furthermore, Seesaw loss~\cite{Wang_2021_CVPR}, SWA~\cite{zhang2020swa} and Soft NMS~\cite{bodla2017soft} are used in the model. A large set of data augmentations is applied during training: Copy-Paste~\cite{kisantal2019augmentation}, rotation, mirroring, cropping, scaling, random brightness, random saturation, random contrast, random color equality, sharpness, random noise, random erasure and local erasure.

Liang et al. show that a relatively simple set of networks and methods can achieve competitive performance, when combined with extensive data augmentation. The Seesaw loss and SWA furthermore both contribute the same 1.1\% AP improvement.


\subsubsection{Jury prize}
Hsu et al. are awarded the jury prize as well as fourth place in the competition. The model used is a HTC~\cite{chen2019hybrid} model with CB-SwinTransformer-Base~\cite{liu2020cbnet,liu2021Swin} backbone. The CB-FPN with Group Normalization~\cite{Wu2018GroupN} is used to "better capture from low to high-level feature representations". Furthermore, the Mask Scoring head~\cite{huang2019mask} replaces the default HTC mask head "to improve model performance on instance’s texture and boundary details". Model Soups~\cite{wortsman2022model} is used to combine models, and SWA~\cite{zhang2020swa} is used to average model weights over multiple training epochs.

\begin{figure}[t]
    \centering
    \includegraphics[width=\linewidth]{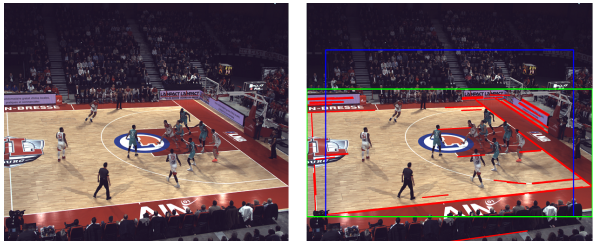}
    \includegraphics[width=\linewidth]{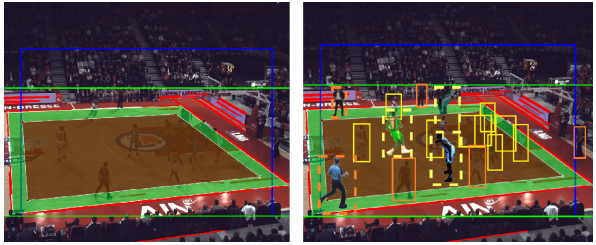}
    \caption{\textit{Basketball Court Detection} method by Hsu et al. The top-left figure is the original image. The top-right one is cropped, with red lines detected by the Canny edge detector and Hough transform. The blue line shows a boundary based on image size, while the green lines indicate dynamic boundary from the detected lines. The bottom-left figure displays a region identified based on the maximum convex hull, which is determined using the endpoints of all lines detected by the Canny-Hough operator. The subclass attributes of the object are determined by its bounding box coordinates. In the bottom-right image, the object marked by a dotted line represents the result of location-based copy-paste augmentation. Figure and caption adapted from technical report by Hsu et al. provided to competition organizers.}
    \label{fig:Hsu-method}
\end{figure}

We award Hsu et al. the jury prize for demonstrating extensive use of (visual) prior knowledge for data-efficient deep learning. First of all, they propose \textit{Basketball Court Detection}, which uses Canny-Hough line detectors~\cite{duda1972use} to find the basketball court and crop the image to contain only the field, as displayed in Fig.~\ref{fig:Hsu-method}. In addition to GridMask augmentation~\cite{chen2020gridmask}, different augmentations are applied to different bounding boxes: RGB curve distortion is applied to those predicted to be "players" specifically to vary skin tone and jersey colors; salt-and-pepper noise and brightness variations are applied to all other boxes, including "referee", "ball" etc. The locations of Copy-Paste augmentations are also changed based on prior information, though exactly how this is done is not clear from the report provided to the organizers. Hsu et al. also choose to implement prior knowledge to save on memory usage, by only inferring the model on the part of the court that is relevant to the task.

The proposed prior-based augmentation pipeline improves AP by 3.7\%. Unfortunately, the resulting model is too resource-intensive for the authors to apply SOTA training settings, which leads to a lacking score on the competition AP metric. However, though the final model by Hsu et al. is not a top performer in this competition, they do improve AP@0.50 by 14.6\% over the 2021 competition winner~\cite{yunusov2021instance} and by 5.9\% over the 2022 competition winner~\cite{yan2022task}, while using less memory and inference time than either competitor.

\begin{table*}[h]
\centering
\caption{Overview of challenge submissions. J indicates jury prize. Bold-faced methods are contributions by the competitors.}
\label{tab:conclusion}
\renewcommand{\arraystretch}{1.4}
\scalebox{0.9}{
\begin{tabular}{@{}lllllc@{}}
\toprule
Rank & Team & Encoder architectures & Data augmentation & Methods & Main metric \\ \midrule

\multicolumn{2}{@{}l}{\textbf{Object detection}} & & & \\ \midrule

1 & \makecell[l]{\textbf{Zhao et al.}} &
    \makecell[l]{Cascade RCNN~\cite{cai2018cascade}, Swin T.~\cite{liu2021Swin}, \\
        ConvNeXt~\cite{liu2022convnet}, ConvNeXtV2~\cite{woo2023convnext}} &
    \makecell[l]{Albumentations~\cite{info11020125}, \\
        PhotoMetricDistortion, \\
        MixUp~\cite{zhang2018mixup}, Auto Augment V2~\cite{cubuk2018autoaugment}} & 
    \makecell[l]{FPN~\cite{lin2017feature}, SWA~\cite{Izmailov2019averaging}, \\
        \textbf{recombined synthetic dataset}, \\
        \textbf{retraining hard classes}} &
    \textbf{34.5} \\

2 \& J & \makecell[l]{Lu et al.} & 
    \makecell[l]{Scaled-YOLOv4~\cite{wang2021scaled}, YOLOv7~\cite{wang2023yolov7}, \\
        YOLOR~\cite{wang2021you}, CBNetv2~\cite{cbnetv2}} & 
    \makecell[l]{Pre-training: random scaling, \\
        random flipping, color jitter; \\
        fine-tuning: Mosaic Augmentation~\cite{bochkovskiy2020yolov4}, \\
        Copy-Paste~\cite{kisantal2019augmentation}, Mix-Up~\cite{zhang2018mixup}, \\
        Cutout~\cite{devries2017improved}} & 
    \makecell[l]{Model Soups~\cite{wortsman2022model}, \\ 
        Weighted Boxes Fusion~\cite{solovyev2021weighted}, \\
        Test-Time Augmentation~\cite{moshkov2020test}, \\
        \textbf{Image Uncertainty Weighted}} &
    33.3 \\

3 & \makecell[l]{Jing et al.} & 
    \makecell[l]{YOLOv7~\cite{yolov7}, YOLOv8x-p2~\cite{githubGitHubUltralyticsultralytics}, \\
        YOLOv8x, YOLOv8x-p6, \\
        Cascade RCNN~\cite{cai2018cascade}} & 
    \makecell[l]{Mosaic Augmentation~\cite{bochkovskiy2020yolov4}, MixUp~\cite{zhang2018mixup}, \\
        Test-Time Augmentation~\cite{moshkov2020test}, \\
        horizontal flip testing} & 
    Weighted Boxes Fusion~\cite{solovyev2021weighted} &
    30.6 \\

4 & \makecell[l]{Wang et al.} & 
    \makecell[l]{YOLOv8~\cite{githubGitHubUltralyticsultralytics}} & 
    \makecell[l]{Mosaic Augmentation~\cite{bochkovskiy2020yolov4}, MixUp~\cite{zhang2018mixup}, \\
        Test-Time Augmentation~\cite{moshkov2020test}, \\
        horizontal flip testing} & 
    \makecell[l]{SparK~\cite{tian2023designing} pre-training, \\
        Weighted Boxes Fusion~\cite{solovyev2021weighted}} &
    30.4 \\

5 & \makecell[l]{Sun et al.} & 
    \makecell[l]{YOLOv8~\cite{githubGitHubUltralyticsultralytics}} & 
    \makecell[l]{HSV, rotation, translation, scaling, \\
        shearing, flipping, \\
        Mosaic Augmentation~\cite{bochkovskiy2020yolov4}, MixUp~\cite{zhang2018mixup}, \\
        Copy-Paste~\cite{kisantal2019augmentation}} & 
    \makecell[l]{GAM~\cite{Zhou2022YOLOv5GEVD}, cross-validation} &
    29.4 \\

6 & \makecell[l]{\textit{Team fha.ddd}} & 
    \makecell[l]{Faster RCNN~\cite{ren2016faster}, \\
        EfficientNet-V2~\cite{tan2021efficientnetv2}} & 
    \makecell[l]{AutoAugment~\cite{cubuk2018autoaugment}} & 
    \makecell[l]{} &
    26.6 \\

\midrule


\multicolumn{2}{@{}l}{\textbf{Instance segmentation}} & & & \\ \midrule

1 & \makecell[l]{
    \textbf{Zhang et al.}} &
    \makecell[l]{Mask RCNN~\cite{he2017mask}, Swin~\cite{liu2021Swin}, \\
        ResNet~\cite{he2016deep}, FPN~\cite{lin2017feature}, \\
        CBNet~\cite{liu2020cbnet}} &
    \makecell[l]{Geometric~\cite{paschali2019manifold}, color space, \\
        sharpness, noise injection, \\
        Copy-Paste~\cite{kisantal2019augmentation}} &
    \makecell[l]{\textbf{Orthogonal Uncertainty} \\
        \textbf{Representation}, \\
        Hybrid Task Cascade~\cite{Chen_2019_CVPR}, \\
        Weighted Boxes Fusion~\cite{solovyev2021weighted}, \\
        Seesaw loss~\cite{Wang_2021_CVPR}, \\
        SWALP~\cite{yang2019swalp}} &
    \textbf{59.0} \\
    
2 & \makecell[l]{
    Lu et al.}&
    \makecell[l]{HTC~\cite{chen2019hybrid}, BEiTv2-L~\cite{peng2022unified}, \\
        ViT-Adapter~\cite{chen2022vision}, Internimage~\cite{wang2023internimage}} &
    \makecell[l]{Mosaic Augmentation~\cite{bochkovskiy2020yolov4}, \\
        Copy-Paste~\cite{kisantal2019augmentation}, Mix-Up~\cite{zhang2018mixup}, \\
        random brightness, random contrast, \\
        random saturation, random scale, \\
        random flip, sharpen and overlay, \\
        blur, Gaussian noise, grid-mask} &
    \makecell[l]{GIoU loss~\cite{rezatofighi2019generalized}, Soft NMS~\cite{bodla2017soft}, \\
        \textbf{expert network} with \\
        SegFormer~\cite{xie2021segformer} and SeMask~\cite{jain2023semask}, \\
        Test-Time Augmentation~\cite{moshkov2020test}, \\
        Model Soups~\cite{wortsman2022model}, \\
        random scaling by \\
        Yunusov et al.~\cite{yunusov2021instance}} &
    58.2 \\
    
3 & \makecell[l]{
    Liang et al.} &
    \makecell[l]{Mask2Former~\cite{cheng2022masked}, YOLOX~\cite{ge2021yolox}, \\
        DETR~\cite{carion2020end}} &
    \makecell[l]{Copy-Paste~\cite{kisantal2019augmentation}, rotation, mirroring, \\
        cropping, scaling, random brightness, \\
        random saturation, random contrast, \\
        random color equality, sharpness, \\
        random noise, random erasure, \\
        local erasure} &
    \makecell[l]{Weighted Boxes Fusion~\cite{solovyev2021weighted}, \\
        Seesaw loss~\cite{Wang_2021_CVPR}, SWA~\cite{zhang2020swa}, Soft NMS~\cite{bodla2017soft}} &
    55.2 \\ 

4 \& J & \makecell[l]{
    Hsu et al.} &
    \makecell[l]{HTC~\cite{chen2019hybrid}, \\
        CB-SwinTransformer-Base~\cite{liu2020cbnet,liu2021Swin}} &
    \makecell[l]{Players: RGB curve distortion; \\
        other objects: salt-and-pepper noise \& \\
        brightness variations; GridMask~\cite{chen2020gridmask}} &
    \makecell[l]{\textbf{Basketball Court Detection}, \\
        GroupNorm~\cite{Wu2018GroupN}, \\
        Mask Scoring R-CNN~\cite{huang2019mask}, \\
        SWA~\cite{zhang2020swa}, Model Soups~\cite{wortsman2022model}} &
    50.9 \\ 


\bottomrule
\end{tabular}
}
\end{table*}

\section{Conclusion}
Table~\ref{tab:conclusion} lists all qualifying entries for each challenge by architecture, data augmentation techniques and any other methods used.

As we organize the VIPriors challenges for the fourth time, we can spot patterns in what makes a successful challenge submission, both for this year and since the first edition. As in all previous editions, heavy use of model ensembling and data augmentation seems to be a recipe for success. Since last year's edition however, Model Soups~\cite{wortsman2022model} and SWA~\cite{zhang2020swa,yang2019swalp} are used rather than standard ensembling. The chosen data augmentation methods also have seen some change, with Mosaic~\cite{bochkovskiy2020yolov4} (especially in combination with YOLO7/YOLO8) and CopyPaste~\cite{kisantal2019augmentation} gaining in popularity. As for the architectures used, there is still a task-dependent bias. YOLO is popular for object detection, even though the winning method uses a Cascade RCNN with ViT-based backbones. For instance segmentation, the picture is more blurred, with both RCNN-based and ViT-based architectures having success.

Encouragingly, we note that novel prior-based methods were more successful in this edition than in the previous edition. Throughout this and previous editions there is a slight signal to indicate that the engineering of prior-based methods does contribute to success in data-efficient deep learning, even though the effort required and the potential increase in inference time and/or memory usage (as in the submission by Hsu et al.) can make it challenging to deploy such methods. However, we must conclude that general deep learning approaches such as increasing model ensembles, heavy data augmentations and extensive tuning are still very effective for data-efficient deep learning. Perhaps we can add a slight metaphorical sweetener to the "bitter lesson"~\cite{sutton2019bitter}, but one has to try hard to taste it.

\Urlmuskip=0mu plus 1mu\relax
\bibliographystyle{splncs04}
\bibliography{main}

%




\end{document}